\documentclass[letterpaper, 10 pt, conference]{ieeeconf}
\IEEEoverridecommandlockouts
\usepackage{amsmath,amssymb,amsfonts}
\usepackage{algorithmic}
\usepackage{graphicx}
\usepackage[export]{adjustbox}
\usepackage{textcomp}
\usepackage{xcolor}
\usepackage{url}
\usepackage{array}
\usepackage{tabularx}
\usepackage{makecell}
\usepackage{booktabs}
\usepackage{multirow}
\usepackage{hyperref}
\usepackage[caption=false]{subfig}
\usepackage{tikz}
\usepackage{pgfplots}
\usepackage{threeparttable}
\newcolumntype{L}{>{\footnotesize}l}
\newcolumntype{M}{>{\small $}l<{$}}
\title{\LARGE \bf
Validation of Human Motor Development Scales with Humanoid Robots} 

\author{Fanxing Meng$^{1}$ and Jing Xiao$^{1}$
\thanks{$^{1}$Robotics Engineering Department,
        Worcester Polytechnic Institute, Worcester, MA, USA
        {\tt\small \{fmeng, jxiao2\}@wpi.edu}}%
}

\begin{document}
\maketitle
\begin{abstract}
We investigate how humanoid robots can learn a series of motor skills defined by human infant motor development scales, and we evaluate which scales are comprehensive enough to serve as self-contained guides for developing gross motor skills up to standing and walking. We represent each motor development scale as a developmental graph that guides the training of a humanoid robot. We then apply this graph within a reinforcement learning framework so that the robot learns skills following the natural phases of human motor development described by the scale. Our work demonstrates that graphs derived from three out of four popular pediatric motor scales can successfully guide humanoid robots to develop these skills. The fourth scale misses some intermediate milestones that the other scales have, which causes some failures. After adding a missing milestone in its graph, it also enables the robot to develop those skills successfully. Video available at \url{https://youtu.be/SR3LUVhohV4}.
 
\end{abstract}

\section{Introduction}

Human infants acquire motor skills in a predictable order: rolling precedes sitting, kneeling precedes crawling, and standing precedes walking. Developmental scientists have captured this progression in standardized motor scales that list milestones and the typical order in which they appear. These scales are used clinically to detect delays in infant development, but they also reflect something deeper: a structured curriculum in which simpler skills provide the physical and coordinative prerequisites for more complex ones.

For humanoid robots, standing up and walking from an arbitrary fallen state remains difficult. Most learning-based approaches either imitate reference trajectories, which must be captured, cleaned, and retargeted for each skill, or require carefully designed reward functions and regularization terms with hand-tuned coefficients. An alternative is to utilize the curriculum that developmental science has already discovered. Our {\em hypothesis} is that if a developmental scale correctly orders human motor learning, it should also guide humanoid robot motor learning, provided the robot's body is sufficiently human-like.

This paper tests that hypothesis. We take four widely used pediatric scales: the Alberta Infant Motor Scale (AIMS)~\cite{AIMS}, the Bayley Scales of Infant and Toddler Development (BSID)~\cite{BSID}, the Gross Motor Function Measure (GMFM)~\cite{GMFM}, and the Peabody Developmental Motor Scales (PDMS)~\cite{PDMS}. We translate each into a developmental graph that directly encodes the milestone sequence and prerequisite structure described in the scale.

We evaluate the four graphs on a simulated 21-DoF humanoid robot in Isaac Lab~\cite{isaaclab}. Three of the four scales produce a consistent developmental progression from supine lying to standing and walking under all four tested training settings. The fourth, PDMS, fails under certain training settings, and notably, it does not include the kneeling-to-standing progression that the other scales use as a bridge to an upright posture. These results validate the idea that human developmental scales can serve as self-contained, science-based curricula for humanoid skill acquisition, and they highlight which milestone sequences matter for standing and walking.

\section{Related Work}

This section introduces the developmental scales used in this research, as well as notable ideas from developmental robotics and curriculum learning.

\subsection{Developmental Scales}

AIMS is a norm-referenced observational assessment of infant gross motor development from birth to independent walking~\cite{AIMS}. It evaluates four positions: supine, prone, sitting, and standing, as well as the transitions between them. AIMS is designed to detect motor delay by comparing an infant's observed skills to age norms.

BSID is a broad assessment of cognitive, language, and motor development in young children~\cite{BSID}. The motor scale includes items from prone control and sitting through crawling, standing, and walking. BSID's progression is similar to that of AIMS but with fewer intermediate milestones, making it a test of whether a sparser developmental graph can still guide a robot to walking.

GMFM was developed to evaluate changes in gross motor function in children with cerebral palsy~\cite{GMFM}. It organizes skills into five dimensions: lying and rolling, sitting, crawling and kneeling, standing, and walking/running/jumping. Because GMFM was built to track functional improvement, its milestones are explicitly ordered by difficulty and prerequisite. This makes it particularly natural to translate into a robot curriculum.

PDMS assesses gross and fine motor skills in children from birth to five years~\cite{PDMS}. Its gross-motor subtest includes reflexes, stationary skills, locomotion, and object manipulation. Unlike AIMS, BSID, and GMFM, the PDMS gross-motor pathway contains only one upright transition (sitting to standing) without the intermediate half-kneeling or high-kneeling milestones that appear in the other scales.

\subsection{Developmental Robotics}

Developmental robotics uses robots to model cognitive and motor development~\cite{devrobbabies}. The field focuses on autonomous, open-ended learning~\cite{devrobsurvey}. It draws on principles such as embodiment (how the body shapes learning)~\cite{embody_emerge}, intrinsic motivation~\cite{intrinsic_motivation}, and the shift from egocentric (self-centered) to object-centric views~\cite{egocentric}. A key concept is Piaget's theory of developmental stages, in which new abilities are built on mastered ones~\cite{Piaget}. This implies that the \textit{sequence} of learning is just as important as the skills themselves. While studies have applied these principles to simpler robots, it remains difficult to create a framework that lets a complex humanoid robot progress through a full trajectory from helplessness to functional behavior.

\subsection{Curriculum Learning}

Curriculum learning structures training from easy to hard to improve efficiency~\cite{curriculum}. In robotics, this takes various forms. Some methods hide difficult goals initially~\cite{CGM} or generate intermediate goals~\cite{HGG}. Others break the learning process into stages~\cite{MultiPhase} or sequence discrete subtasks~\cite{AbstractAdaptive,PlanHierarchicallyFromCurriculum}. A third category changes the environment, such as starting the agent in states progressively farther from the goal~\cite{VaPRL,SGC} or slowly increasing terrain difficulty~\cite{SixLegCL}. The main goal of these techniques is usually to speed up convergence for a specific, single task.

\section{Method}

Our method converts each scale into a developmental graph that guides a reinforcement learning (RL) strategy for a humanoid robot to acquire related motor skills.

\subsection{Developmental Graphs}

\begin{figure}[tbp]
    \centering
    \subfloat[AIMS]{\includegraphics[width=0.26\textwidth]{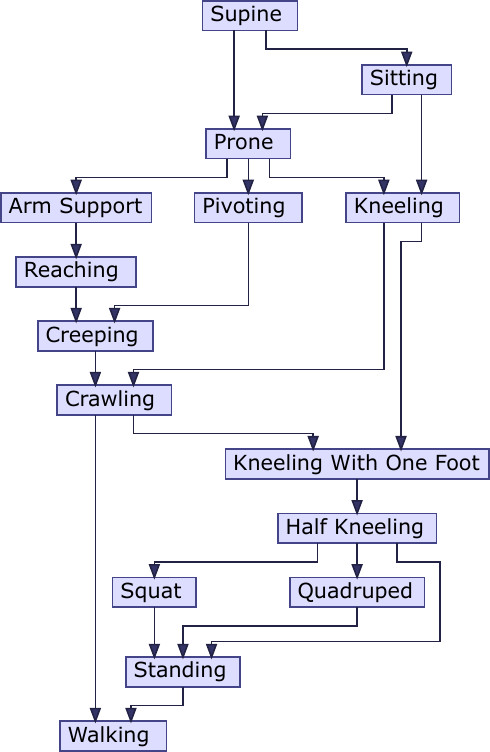}}\hfill
    \subfloat[BSID]{\includegraphics[width=0.192\textwidth]{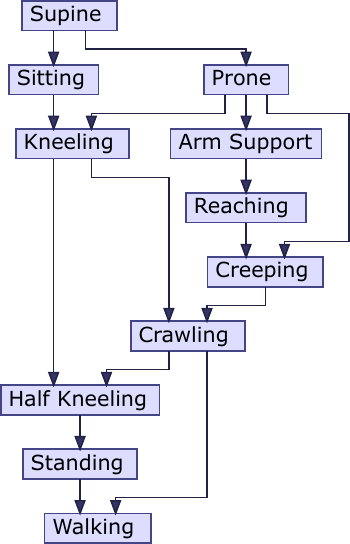}}\hfill
    \subfloat[GMFM]{\includegraphics[width=0.35\textwidth]{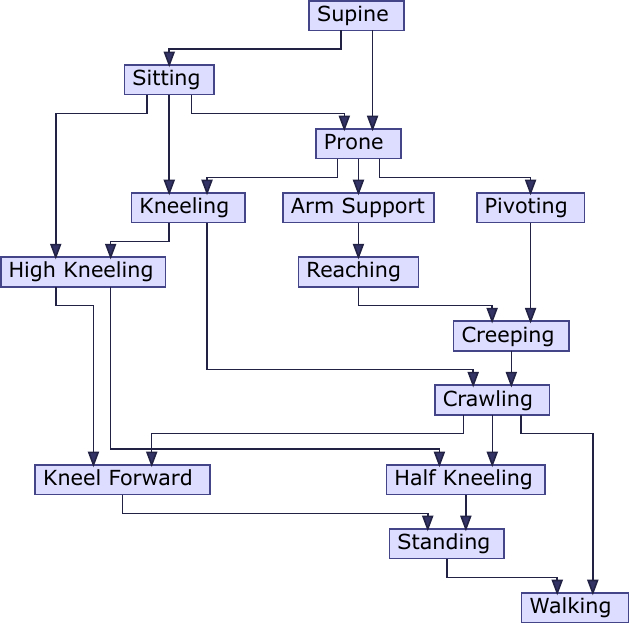}}\hfill
    \subfloat[PDMS]{\includegraphics[width=0.25\textwidth]{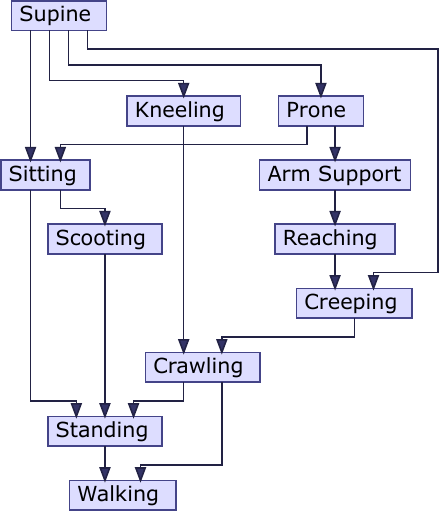}}\hfill
    \caption{Master graphs derived from each developmental scale. Nodes are topologically sorted; each horizontal row of nodes represents one topological level of the graph. Solid arrow lines indicate that prerequisite milestones need only be achieved once in an episode to activate subsequent nodes.}
    \label{master_graphs}
\end{figure}

Each pediatric scale is translated into a developmental graph, as shown in Fig.~\ref{master_graphs}. The master graph contains one node per gross milestone (e.g., prone, sitting, kneeling, standing, walking).\footnote{Milestones are selected only if they do not rely on external objects or require physical interaction with an examiner.} Directed edges represent prerequisite conditions or pose transitions defined by the scale.\footnote{When a scale contains two-way transitions between two poses, such as rolling over from supine to prone and rolling back from prone to supine, the later-acquired transition (rolling back) is removed to keep the graph acyclic.}

We implement the developmental graph as an achievement-gating reward structure: a reward node can contribute to the total reward only after at least one of its parent nodes has been achieved once in the current episode. In RL terms, a node is activated when at least one parent's reward value exceeds a passing score, even briefly.

\begin{figure}[tbp]
     \smallskip
     \smallskip
     \smallskip
    \centering
    \subfloat[Kneeling with one foot]{\includegraphics[width=0.175\textwidth]{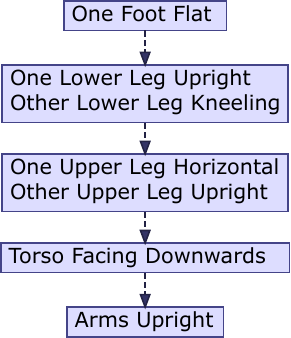}}\hfill
    \subfloat[Half kneeling]{\includegraphics[width=0.175\textwidth]{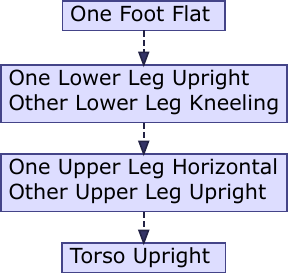}}\hfill
    \subfloat[High kneeling]{\includegraphics[width=0.15\textwidth]{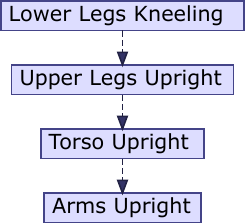}}\hfill
    \subfloat[Kneel forward]{\includegraphics[width=0.15\textwidth]{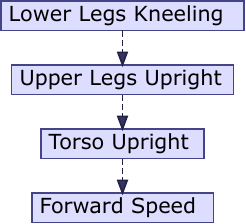}}\hfill
    \subfloat[Quadruped]{\includegraphics[width=0.14\textwidth]{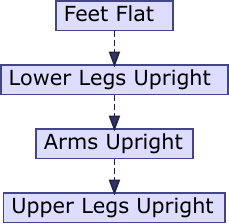}}\hfill
    \subfloat[Squat]{\includegraphics[width=0.18\textwidth]{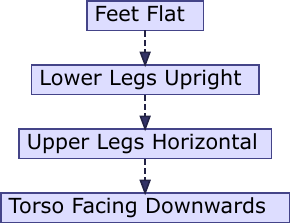}}\hfill
    \subfloat[Standing]{\includegraphics[width=0.142\textwidth]{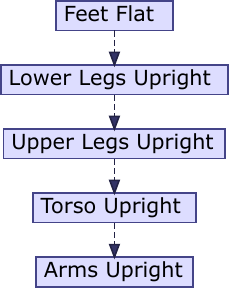}}\hfill
    \subfloat[Walking]{\includegraphics[width=0.11\textwidth]{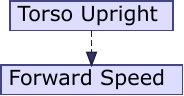}}\hfill
    \caption{Mini-graph structures used inside the master graphs. Dashed arrow lines indicate that prerequisite milestones must remain achieved to activate subsequent nodes.}
    \label{mini_graphs}
\end{figure}

Because upright milestones are mechanically composite, we further decompose them into mini-graphs, as shown in Fig.~\ref{mini_graphs}. For example, standing is represented as a sequence of force-propagation conditions: both feet must be flat on the ground, both lower legs must be upright, both upper legs must be upright, the torso must be upright, and the arms should be brought in. Each condition is a node in the standing mini-graph, and the conditions must be satisfied in order. A node in a mini-graph is activated only when all of its internal predecessors are simultaneously above their passing scores; this forces the robot to maintain, not merely visit, the supporting posture.

Formally, let $r_{i,t}$ be the instantaneous raw reward for node $i$ at time $t$, normalized between $-1$ and $1$. For a transition from node $i$ to node $j$, we define a passing score $p_{i,j}$ and an achievement score
\begin{equation}
a_{i,j,t} =
\begin{cases}
\bigl(r_{i,t} - p_{i,j}\bigr)^+ & \text{if $j$ requires $i$ to persist,} \\[4pt]
\bigl(\max\limits_{t^*\le t} r_{i,t^*} - p_{i,j}\bigr)^+ & \text{otherwise,}
\end{cases}
\end{equation}
where $(\square)^+ = \max\{0, \square\}$.

The total reward is then
\begin{equation}
R_t = \sum_j \sum_{i\in\mathcal{P}(j)} a_{i,j,t}\, m_j\, \bigl(r_{j,t} + b_j\bigr),
\end{equation}
where $\mathcal{P}(j)$ is the set of parent nodes of $j$, $m_j$ is a level-dependent multiplier that encourages the robot to learn more advanced milestones, and $b_j$ is a bias automatically calculated at runtime that shifts the raw reward so that the activated node evaluates to a positive value.

The multiplier $m_j$ is assigned automatically from the graph topology. For nodes in the master graph, the multiplier grows quadratically with the longest-path level from the start node. For nodes inside a mini-graph, the first node inherits the master-graph multiplier of its parent milestone, and subsequent nodes receive $m_{\text{first}} \sqrt{\ell}$, where $\ell$ is the internal level. This gives the initial supporting condition a large enough weight while still increasing the weight of later posture refinements, but it avoids shadowing the quadratic growth of other nodes in the master graph. No reward weights are tuned by hand. The geometric definitions of the milestones are translated directly from the scale descriptions.

\subsection{Reward Functions}

\begin{figure*}[tbp]
    \centering
    \subfloat[Reaching]{\includegraphics[width=0.25\textwidth]{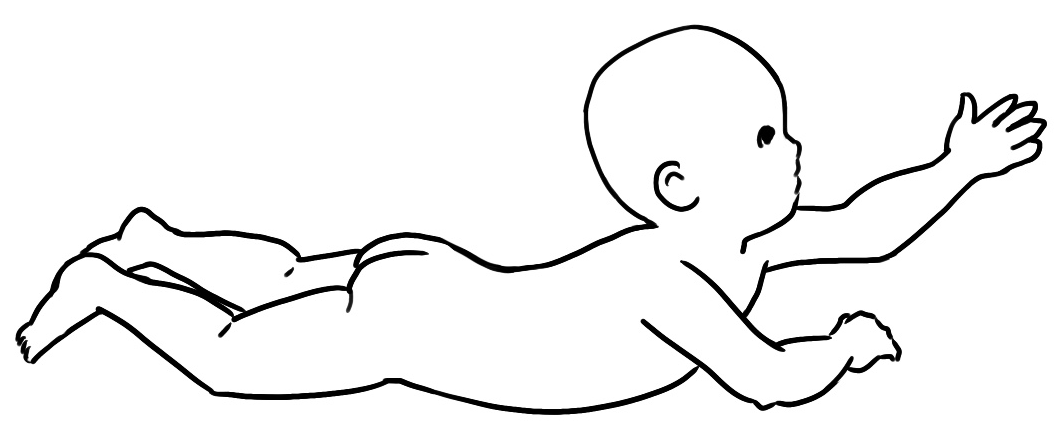}}\hfill
    \subfloat[Kneeling with one foot]{\includegraphics[width=0.17\textwidth]{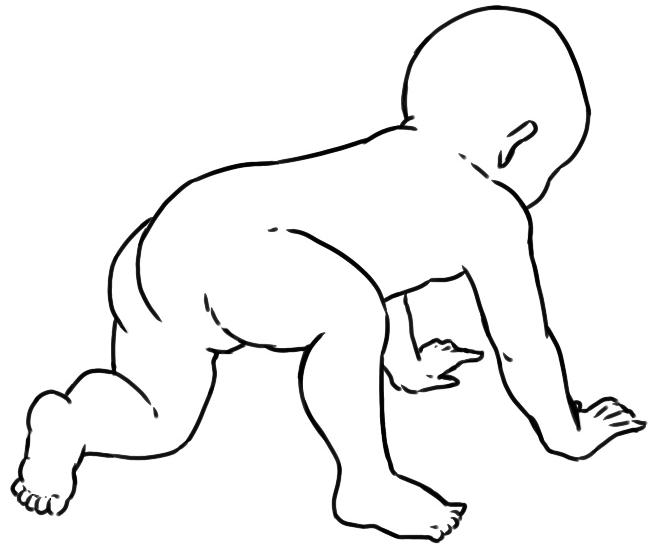}}\hfill
    \subfloat[Half kneeling]{\includegraphics[width=0.11\textwidth]{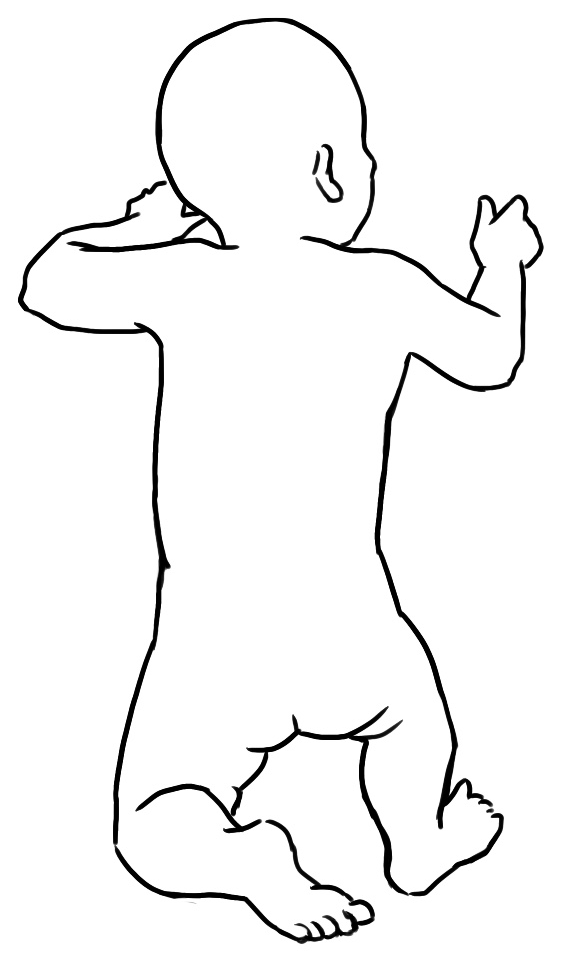}}\hfill
    \subfloat[Squat]{\includegraphics[width=0.14\textwidth]{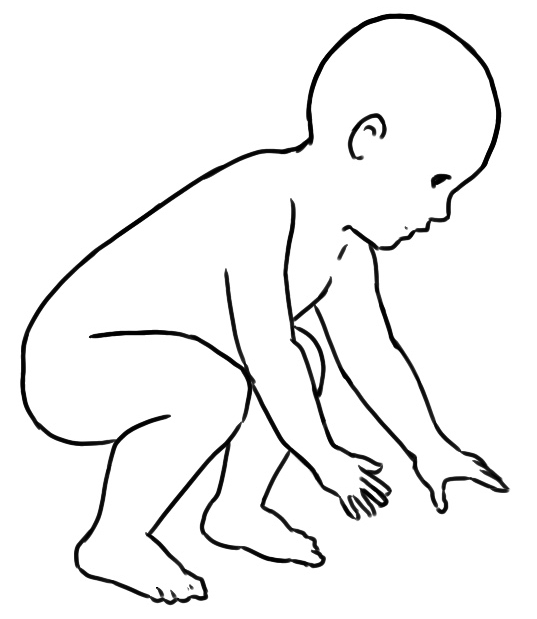}}\hfill
    \subfloat[Quadruped]{\includegraphics[width=0.15\textwidth]{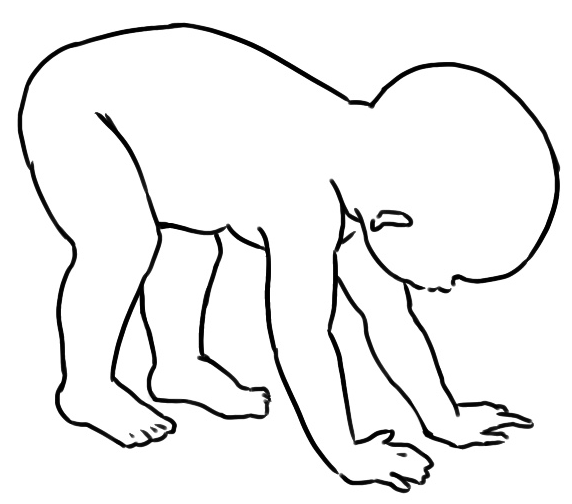}}\hfill
    \caption{Reproduced examples of visually represented pose milestones in AIMS  from ~\cite{AIMS}.}
    \label{pose_graphs}
\end{figure*}

\begin{figure}[tbp]
    \centering
    \includegraphics[width=0.45\textwidth]{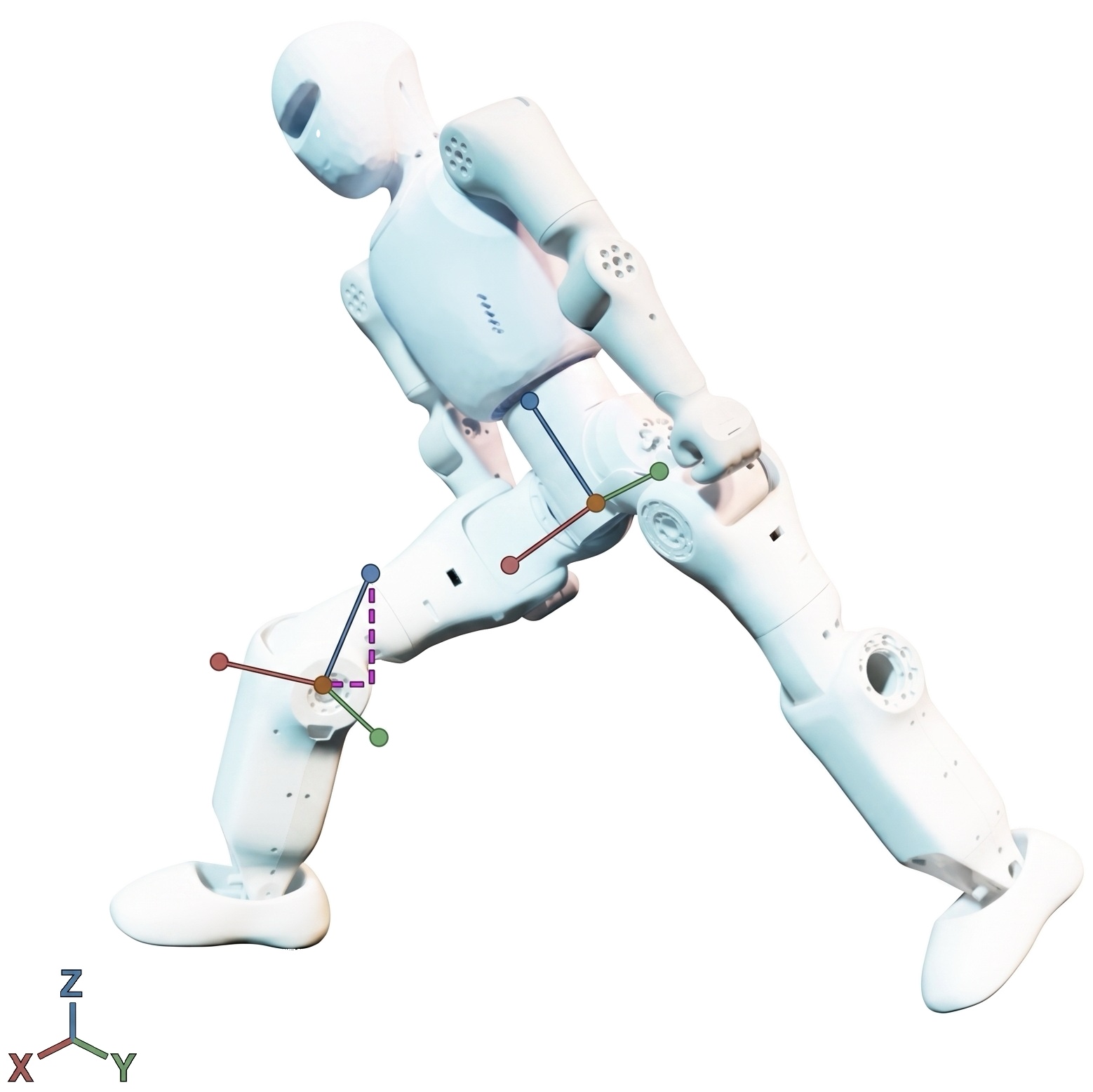}
    \caption{Coordinate systems used in the reward definitions. The world frame is drawn on the lower left, with the vertical $z$-axis aligned with gravity. The robot's root reference frame is attached at the pelvis. A link-local frame is shown on the lower leg; its local $\hat{z}$ unit vector (superior direction) projection (purple dashed line) onto the world $z$-axis is used to compute orientation-based reward term $\hat{z}_{\textrm{lower leg}'_z}$.}
    \label{fig:coordinate_systems}
\end{figure}

The reward definition for each node is based on the poses graphed in AIMS; a few drawings are reproduced in Fig.~\ref{pose_graphs} for illustration. For static poses, each reward term is the cosine similarity between the depicted orientation of a body part and its actual orientation, so perfect alignment yields $+1$ and the opposite orientation yields $-1$. 
 
For dynamic motions, the reward term is the robot's velocity along the relevant torso axis described in the scale: for example, crawling and creeping are rewarded along the torso's superior axis, while walking and scooting are rewarded along the torso's anterior axis.

The same canonical pose definitions are reused when a milestone appears in more than one scale. Fig.~\ref{fig:coordinate_systems} illustrates the relationship between the world frame, the root reference frame, and a link-local frame for computing orientation-based rewards. All reward component definitions are listed in Tables~\ref{rewards} and \ref{components}.

\begin{table}[tbp]
     \smallskip
     \smallskip
     \smallskip
\centering
\caption{Definition of standalone node rewards.}
\label{rewards}
\renewcommand{\arraystretch}{1.1}
\begin{tabular}{L M}
\toprule
\textbf{Description} & \textbf{Expression} \\
\midrule
\multicolumn{2}{l}{\textbf{Prone}} \\
\quad Torso's anterior axis points downward & -\hat{x}_{\textrm{torso}_z} \\

\multicolumn{2}{l}{\textbf{Arm support}} \\
\quad Arms' superior axes point upward & \hat{z}_{\textrm{arms}_z} \\

\multicolumn{2}{l}{\textbf{Pivoting}} \\
\quad Torso rotates horizontally & \omega_{\textrm{root}_z} \\

\multicolumn{2}{l}{\textbf{Reaching}} \\
\quad Arms' inferior axes along torso's superior axis & -\hat{z}_{\textrm{arms}}\cdot\hat{z}_{\textrm{torso}} \\

\multicolumn{2}{l}{\textbf{Creeping}} \\
\quad Root velocity along torso's superior axis & \vec{v}_\textrm{root} \cdot \hat{z}_\textrm{torso} \\

\multicolumn{2}{l}{\textbf{Sitting}} \\
\quad Torso's superior axis points upward & \hat{z}_{\textrm{torso}_z} \\

\multicolumn{2}{l}{\textbf{Scooting}} \\
\quad Torso's superior axis points upward & \hat{z}_{\textrm{torso}_z} \\
\quad Root velocity along its anterior axis & \vec{v}_\textrm{root} \cdot \hat{x}_\textrm{root} \\

\multicolumn{2}{l}{\textbf{Kneeling}} \\
\quad Arms' superior axes point upward & \hat{z}_{\textrm{arms}_z} \\
\quad Upper legs' superior axes point upward & \hat{z}_{\textrm{upper legs}_z} \\
\quad Lower legs' superior axes along torso's & \hat{z}_\textrm{lower legs} \cdot \hat{z}_\textrm{torso} \\

\multicolumn{2}{l}{\textbf{Crawling}} \\
\quad Arms' superior axes point upward & \hat{z}_{\textrm{arms}_z} \\
\quad Upper legs' superior axes point upward & \hat{z}_{\textrm{upper legs}_z} \\
\quad Lower legs' superior axes along torso's & \hat{z}_\textrm{lower legs} \cdot \hat{z}_\textrm{torso} \\
\quad Root velocity along torso's superior axis & \vec{v}_\textrm{root} \cdot \hat{z}_\textrm{torso} \\
\bottomrule
\end{tabular}
\begin{tablenotes}
    \small
    \item Anatomical terms (`anterior', `superior') refer to local link axes.
    \item Spatial terms (`upward', `downward') refer to world axes.
    \item Unit vectors are along local link axes ($\hat{x}$: anterior, $\hat{y}$: left, $\hat{z}$: superior, when standing at attention).
    \item The second subscript denotes the component along a world axis.
    \item ``$\cdot$'' denotes the vector dot product used to measure collinearity.
\end{tablenotes}
\end{table}

\begin{table}[tbp]
\centering
\caption{Definition of mini-graph reward components.}
\renewcommand{\arraystretch}{1.15}
\label{components}
\begin{tabular}{L M}
\toprule
\textbf{Description} & \textbf{Expression} \\
\midrule
\multicolumn{2}{l}{\textbf{Kneeling with one foot}} \\
\quad One foot flat & \hat{z}_{\textrm{foot}'_z} + \hat{y}_{\textrm{root}}\cdot\hat{y}_{\textrm{foot}'} \\
\quad One lower leg upright &  \hat{z}_{\textrm{lower leg}'_z} \\
\quad Other lower leg kneeling & \hat{y}_{\textrm{lower leg}''}\cdot\hat{y}_\textrm{root} - \hat{x}_{\textrm{lower leg}''_z} \\
\quad One upper leg horizontal & \hat{x}_{\textrm{upper leg}'_z} \\
\quad Other upper leg upright & \hat{z}_{\textrm{upper leg}''_z} \\
\quad Torso faces downward & -\hat{x}_{\textrm{torso}_z} \\
\quad Arms upright & \hat{z}_{\textrm{arms}_z} \\

\multicolumn{2}{l}{\textbf{High kneeling}} \\
\quad Lower legs kneeling & \hat{y}_{\textrm{lower legs}}\cdot\hat{y}_\textrm{root} - \hat{x}_{\textrm{lower legs}_z} \\
\quad Upper legs upright & \hat{z}_{\textrm{upper legs}_z} \\
\quad Torso upright & \hat{z}_{\textrm{torso}_z} + \hat{x}_{\textrm{root}}\cdot\hat{x}_{\textrm{torso}} \\
\quad Arms upright & \hat{z}_{\textrm{arms}_z} \\

\multicolumn{2}{l}{\textbf{Kneel forward}} \\
\quad Lower legs kneeling & \hat{y}_{\textrm{lower legs}}\cdot\hat{y}_\textrm{root} - \hat{x}_{\textrm{lower legs}_z} \\
\quad Upper legs upright & \hat{z}_{\textrm{upper legs}_z} \\
\quad Torso upright & \hat{z}_{\textrm{torso}_z} + \hat{x}_{\textrm{root}}\cdot\hat{x}_{\textrm{torso}} \\
\quad Forward speed & \vec{v}_\textrm{root} \cdot \hat{x}_\textrm{root} \\

\multicolumn{2}{l}{\textbf{Half kneeling}} \\
\quad One foot flat & \hat{z}_{\textrm{foot}'_z} + \hat{y}_{\textrm{root}}\cdot\hat{y}_{\textrm{foot}'} \\
\quad One lower leg upright &  \hat{z}_{\textrm{lower leg}'_z} \\
\quad Other lower leg kneeling & \hat{y}_{\textrm{lower leg}''}\cdot\hat{y}_\textrm{root} -\hat{x}_{\textrm{lower leg}''_z} \\
\quad One upper leg horizontal & \hat{x}_{\textrm{upper leg}'_z} \\
\quad Other upper leg upright & \hat{z}_{\textrm{upper leg}''_z} \\
\quad Torso upright & \hat{z}_{\textrm{torso}_z} \\

\multicolumn{2}{l}{\textbf{Quadruped}} \\
\quad Feet flat & \hat{z}_{\textrm{feet}_z} \\
\quad Lower legs upright & \hat{z}_{\textrm{lower legs}_z} \\
\quad Arms upright & \hat{z}_{\textrm{arms}_z} \\
\quad Upper legs upright & \hat{z}_{\textrm{upper legs}_z} \\

\multicolumn{2}{l}{\textbf{Squat}} \\
\quad Feet flat & \hat{z}_{\textrm{feet}_z} \\
\quad Lower legs upright & \hat{z}_{\textrm{lower legs}_z} \\
\quad Upper legs horizontal & \hat{x}_{\textrm{upper legs}_z} \\
\quad Torso faces downward & -\hat{x}_{\textrm{torso}_z} \\

\multicolumn{2}{l}{\textbf{Standing}} \\
\quad Feet flat & \hat{z}_{\textrm{feet}_z} + \hat{x}_{\textrm{root}}\cdot\hat{x}_{\textrm{feet}} \\
\quad Lower legs upright & \hat{z}_{\textrm{lower legs}_z} \\
\quad Upper legs upright & \hat{z}_{\textrm{upper legs}_z} \\
\quad Torso upright & \hat{z}_{\textrm{torso}_z} + \hat{x}_{\textrm{root}}\cdot\hat{x}_{\textrm{torso}} \\
\quad Arms upright & \hat{z}_{\textrm{arms}_z} \\

\multicolumn{2}{l}{\textbf{Walking}} \\
\quad Torso upright & \hat{z}_{\textrm{torso}_z} + \hat{x}_{\textrm{root}}\cdot\hat{x}_{\textrm{torso}} \\
\quad Forward speed & \vec{v}_\textrm{root}\cdot\hat{x}_\textrm{root} \\
\bottomrule
\end{tabular}
\begin{tablenotes}
    \small
    \item $'$ denotes either a left or right limb; $''$ denotes the other limb.
\end{tablenotes}
\end{table}
\section{Experiments}

We implemented the four developmental graphs in Isaac Lab. The robot is a 21-DoF humanoid model based on the Noetix Bumi robot, with torque-controlled joints. Training uses 4,096 parallel environments on a single consumer-grade GPU (less than 8~GB of VRAM), with each run limited to 1,250 episodes and taking less than 12 GPU-hours. Each episode lasts 10 seconds at a 50~Hz control frequency. The policy is a feed-forward multi-layer perceptron (MLP) with layers $[800, 400, 200, 100]$ trained with proximal policy optimization (PPO)~\cite{PPO}. Domain randomization includes mass scaling, friction scaling, observation noise, action noise, and periodic gravity perturbations. Table~\ref{hyper_parameters} lists the simulation and randomization hyper-parameters.

\begin{table}[tbp]
     \smallskip
     \smallskip
     \smallskip
\centering
\begin{threeparttable}
\renewcommand{\arraystretch}{1.1}
\caption{List of hyper-parameters.}
\label{hyper_parameters}
\centering
\begin{tabular}{@{}l l | l l@{}}
\toprule
\textbf{Simulation} & & \textbf{Randomization} & \\
    \cmidrule(lr){1-2}\cmidrule(lr){3-4}
Simulation timestep & 0.005 s & Observations & $+\mathcal{N}(0, 0.002)$ \\
Position iterations & 8 & Actions & $+\mathcal{N}(0, 0.05)$ \\
Velocity iterations & 4 & Gravity & $+\mathcal{N}(0, 0.2)$ \\
Contact offset & 0.01 m & Body mass & $\times \mathcal{U}\left[0.8, 1.2\right]$ \\
Rest offset & 0.0 m & Friction & $\times \mathcal{U}\left[0.7, 1.3\right]$ \\
Bounce threshold & 0.5 m/s & Damping \& stiffness & $\times \mathcal{U}\left[0.8, 1.2\right]$ \\
\bottomrule
\vspace{-10pt}
\end{tabular}
\begin{tablenotes}
    \small
    \item $+\mathcal{N}(\textrm{mean}, \textrm{variance})$ denotes additive normal noise.
    \item $\times \mathcal{U}\left[\textrm{lower}, \textrm{upper}\right]$ denotes multiplicative uniform scaling.
\end{tablenotes}
\end{threeparttable}
\end{table}

The observation vector is 419-dimensional and consists of root height, root orientation, root linear and angular velocity, positions/orientations/velocities of all 22 links, scaled joint positions and velocities, previous actions, previous torque commands, sinusoidal signals, and episode progress. The action vector is 21-dimensional and represents normalized joint torque commands which are multiplied by each joint's maximum torque and the current torque scale before being applied.

Besides the developmental graph, our training setup uses three implementation choices borrowed from prior work in developmental and reinforcement learning~\cite{TimeChamber,PeriodicRewardComposition,sata}. First, the positions, orientations, and velocities of the robot links are expressed relative to the robot's own root frame affixed to the pelvis. Second, the observation includes sixteen time-varying sinusoids to mimic the effect of central pattern generator signals. Third, the available joint torque is scaled linearly from zero at the start of each episode to the full motor limit over the first half of the episode. These are training aids taken from earlier studies; the developmental graph is the only component studied in this paper.

We trained each developmental graph at four passing scores (50\%, 60\%, 70\%, and 80\%). A rollout was considered successful for standing if the robot maintained a torso-upright posture with head height above a threshold for at least 100 consecutive steps, and successful for walking if its forward speed exceeded 0.5~m/s.

\begin{figure*}[tbp]
\centering
\subfloat[AIMS]{%
\includegraphics[width=0.093\textwidth]{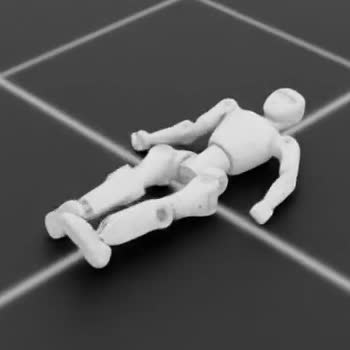}
\hfill
\includegraphics[width=0.093\textwidth]{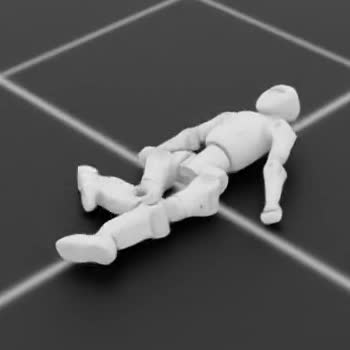}
\hfill
\includegraphics[width=0.093\textwidth]{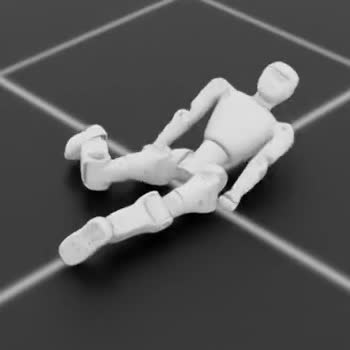}
\hfill
\includegraphics[width=0.093\textwidth]{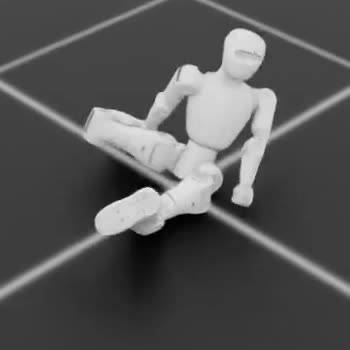}
\hfill
\includegraphics[width=0.093\textwidth]{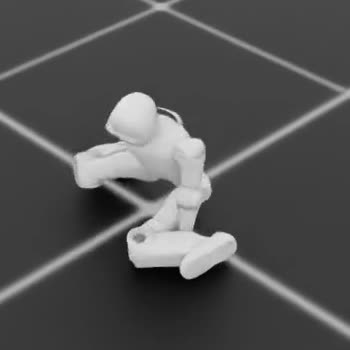}
\hfill
\includegraphics[width=0.093\textwidth]{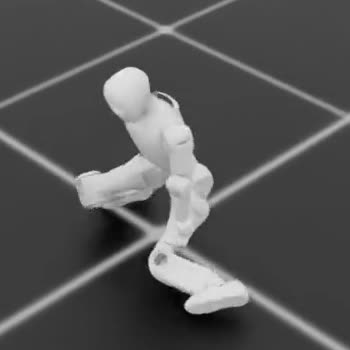}
\hfill
\includegraphics[width=0.093\textwidth]{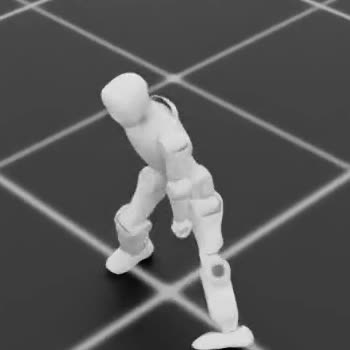}
\hfill
\includegraphics[width=0.093\textwidth]{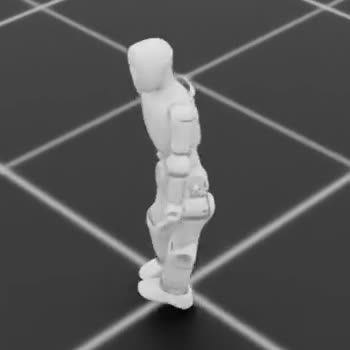}
\hfill
\includegraphics[width=0.093\textwidth]{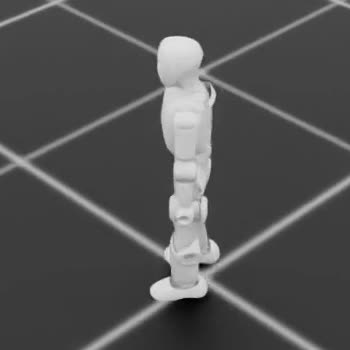}
\hfill
\includegraphics[width=0.093\textwidth]{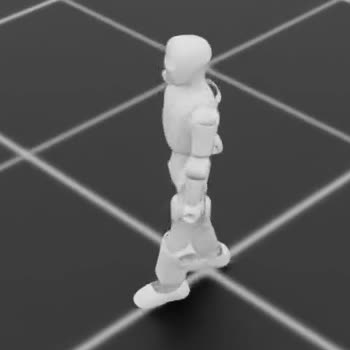}
}

\subfloat[BSID]{%
\includegraphics[width=0.093\textwidth]{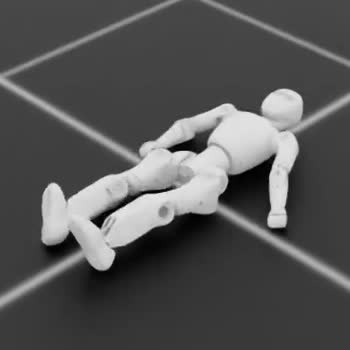}
\hfill
\includegraphics[width=0.093\textwidth]{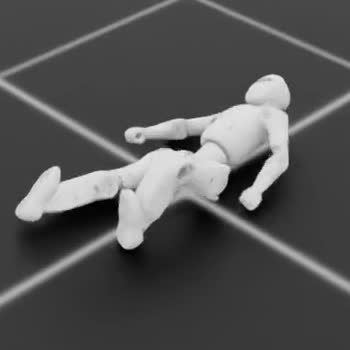}
\hfill
\includegraphics[width=0.093\textwidth]{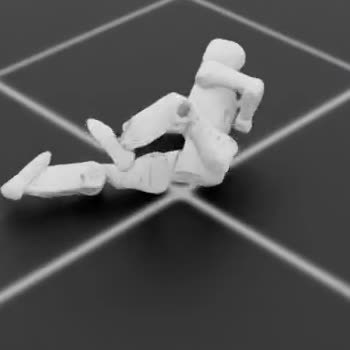}
\hfill
\includegraphics[width=0.093\textwidth]{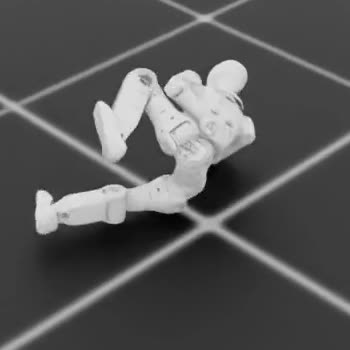}
\hfill
\includegraphics[width=0.093\textwidth]{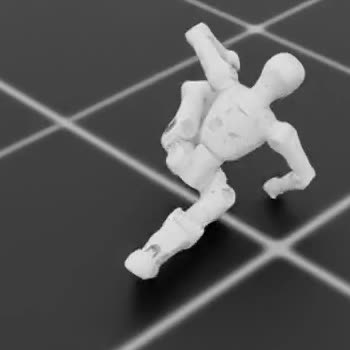}
\hfill
\includegraphics[width=0.093\textwidth]{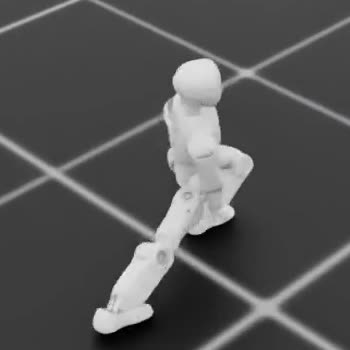}
\hfill
\includegraphics[width=0.093\textwidth]{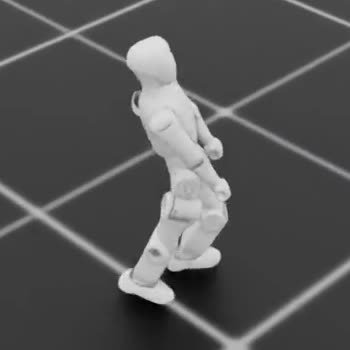}
\hfill
\includegraphics[width=0.093\textwidth]{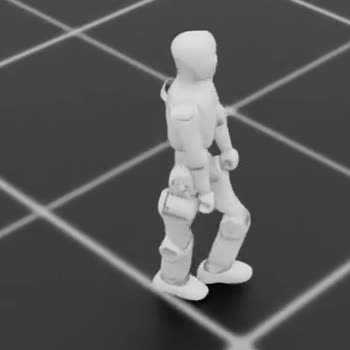}
\hfill
\includegraphics[width=0.093\textwidth]{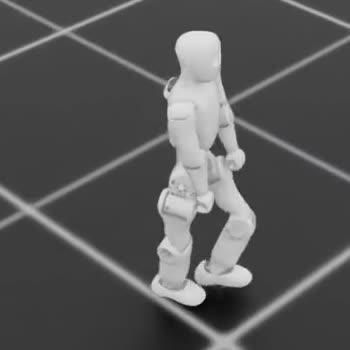}
\hfill
\includegraphics[width=0.093\textwidth]{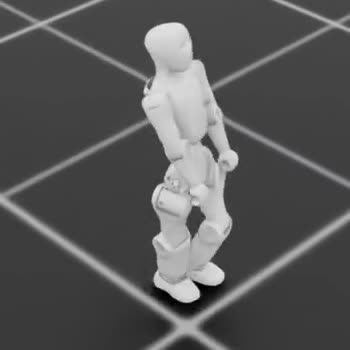}
}

\subfloat[GMFM]{%
\includegraphics[width=0.093\textwidth]{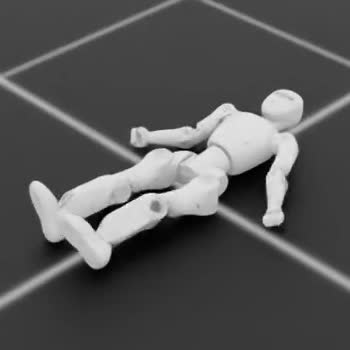}
\hfill
\includegraphics[width=0.093\textwidth]{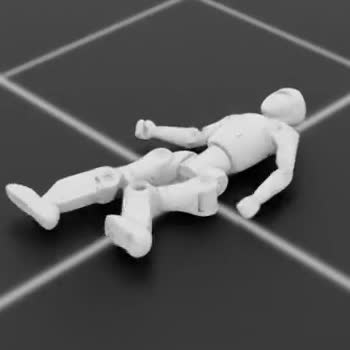}
\hfill
\includegraphics[width=0.093\textwidth]{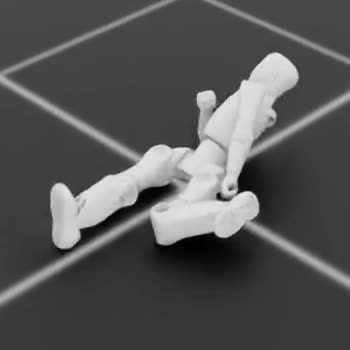}
\hfill
\includegraphics[width=0.093\textwidth]{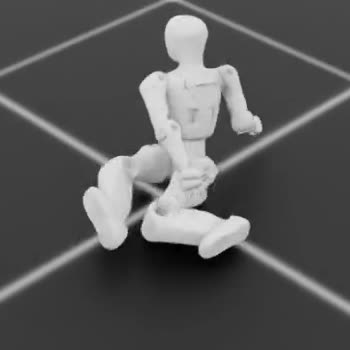}
\hfill
\includegraphics[width=0.093\textwidth]{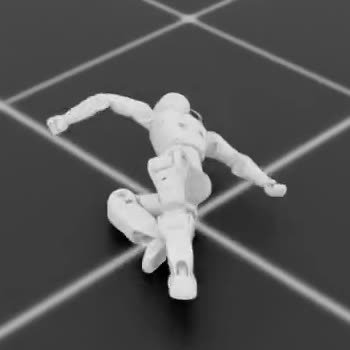}
\hfill
\includegraphics[width=0.093\textwidth]{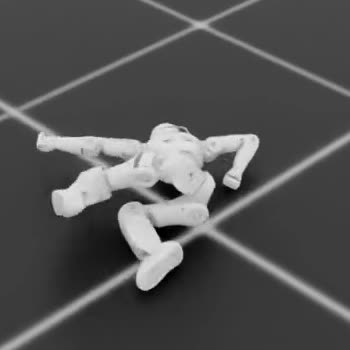}
\hfill
\includegraphics[width=0.093\textwidth]{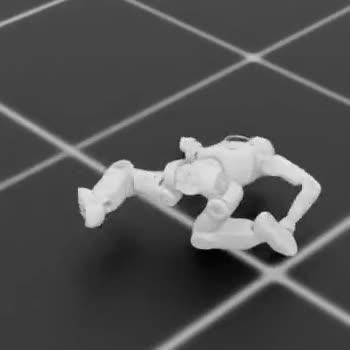}
\hfill
\includegraphics[width=0.093\textwidth]{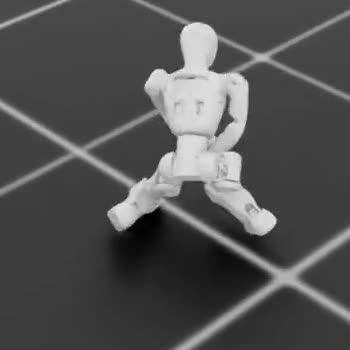}
\hfill
\includegraphics[width=0.093\textwidth]{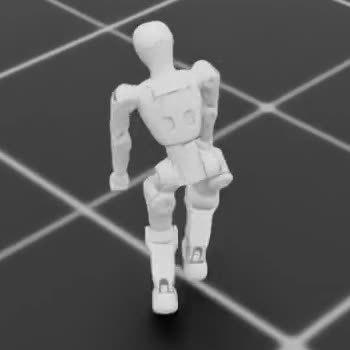}
\hfill
\includegraphics[width=0.093\textwidth]{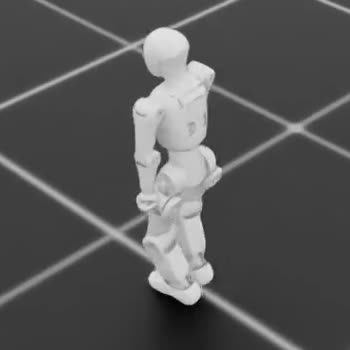}
}

\subfloat[PDMS]{%
\includegraphics[width=0.093\textwidth]{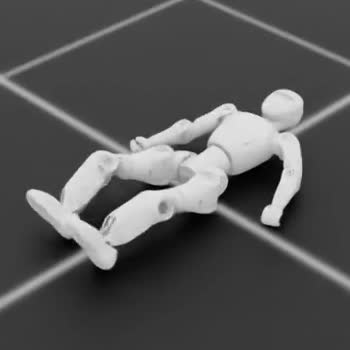}
\hfill
\includegraphics[width=0.093\textwidth]{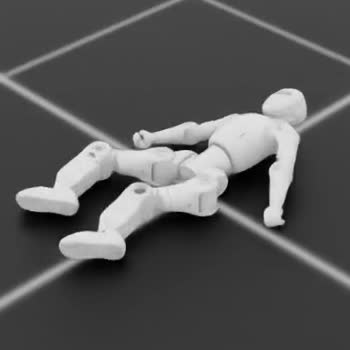}
\hfill
\includegraphics[width=0.093\textwidth]{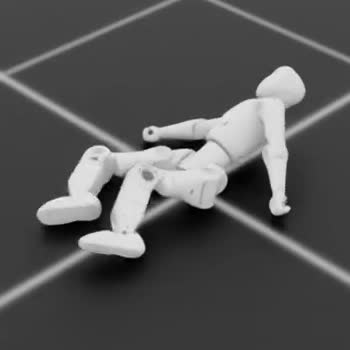}
\hfill
\includegraphics[width=0.093\textwidth]{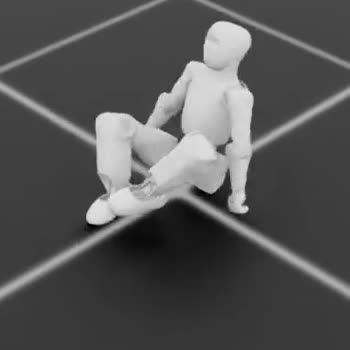}
\hfill
\includegraphics[width=0.093\textwidth]{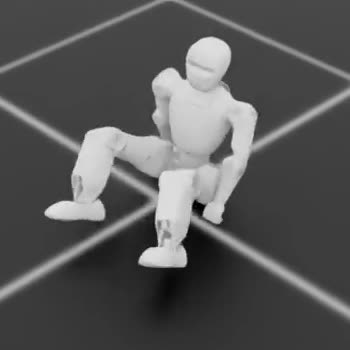}
\hfill
\includegraphics[width=0.093\textwidth]{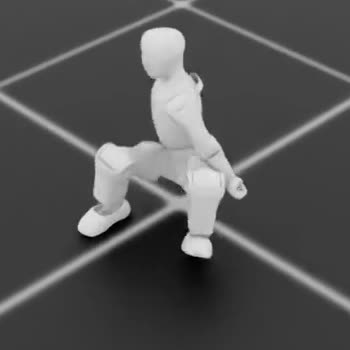}
\hfill
\includegraphics[width=0.093\textwidth]{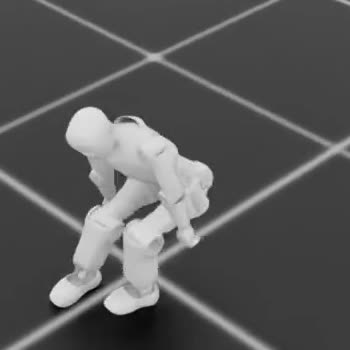}
\hfill
\includegraphics[width=0.093\textwidth]{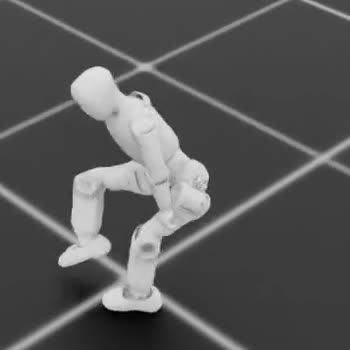}
\hfill
\includegraphics[width=0.093\textwidth]{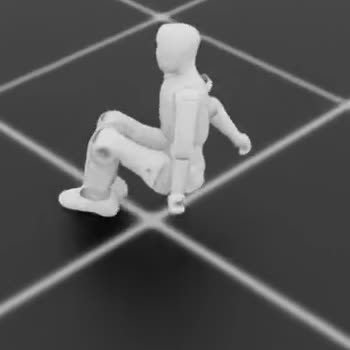}
\hfill
\includegraphics[width=0.093\textwidth]{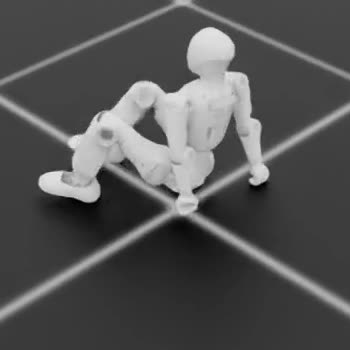}
}

\caption{Standing-up sequence of robots trained using the four developmental graphs with a 50\% passing score. PDMS fails to stand up under this training setting.}
\label{fig:standup}
\end{figure*}

\section{Results}

Robots trained under the AIMS, BSID, and GMFM graphs learned the full progression from supine lying to standing and walking under all four tested passing-score settings. Fig.~\ref{fig:standup} shows representative snapshots of the learned standing-up motion for each graph at the 50\% setting.

The PDMS graph did not learn sustained standing at the 50\% setting or at the 80\% setting. This is consistent with the structure of the PDMS gross-motor pathway: it defines a direct sitting-to-standing transition but omits the half-kneeling and high-kneeling intermediates that appear in AIMS, BSID, and GMFM. Without those intermediate support postures, the reward landscape contains a large mechanical gap that the policy could not bridge, and the robot falls back to a scooting pose.

To test whether this structural difference explains the failure, we inserted the half-kneeling milestone, together with its mini-graph, between the kneeling and standing nodes of the PDMS graph and retrained at all four passing scores, leaving every other setting unchanged. The augmented graph learned to stand and walk at the 50\% passing score, where the original PDMS graph failed. It still did not succeed at the 60\%, 70\%, or 80\% settings. This test supports the explanation that the kneeling bridge is an important structural feature for getting up from the ground. At the same time, it shows that adding a single milestone does not by itself make a sparse curriculum robust across all passing-score settings.

Table~\ref{tab:success_rates} reports the success rate of standing without falling and the final walking speed. Table~\ref{tab:graph_comparison} compares the time-to-skill statistics of the four graphs under different passing-score settings. Success rates do not simply increase with passing score. Under domain randomization, some robots in the population have limb masses or action noise that make it hard to reach the near-perfect orientations required by a high passing score. When a prerequisite milestone does not beat the threshold, downstream milestones are never unlocked, so those robots cannot progress. This is why the highest passing score sometimes gives a lower success rate than a moderate one, and why some cells in Table~\ref{tab:graph_comparison} are empty at higher passing scores even though the same milestone was learned at a lower threshold. This also shows that some parts of skill development are overlapping or concurrent in practice: setting a passing score too high can prevent the robot from developing related skills together.

\begin{figure*}[tbp]
     \smallskip
     \smallskip
    \centering
    \includegraphics[width=0.9\textwidth]{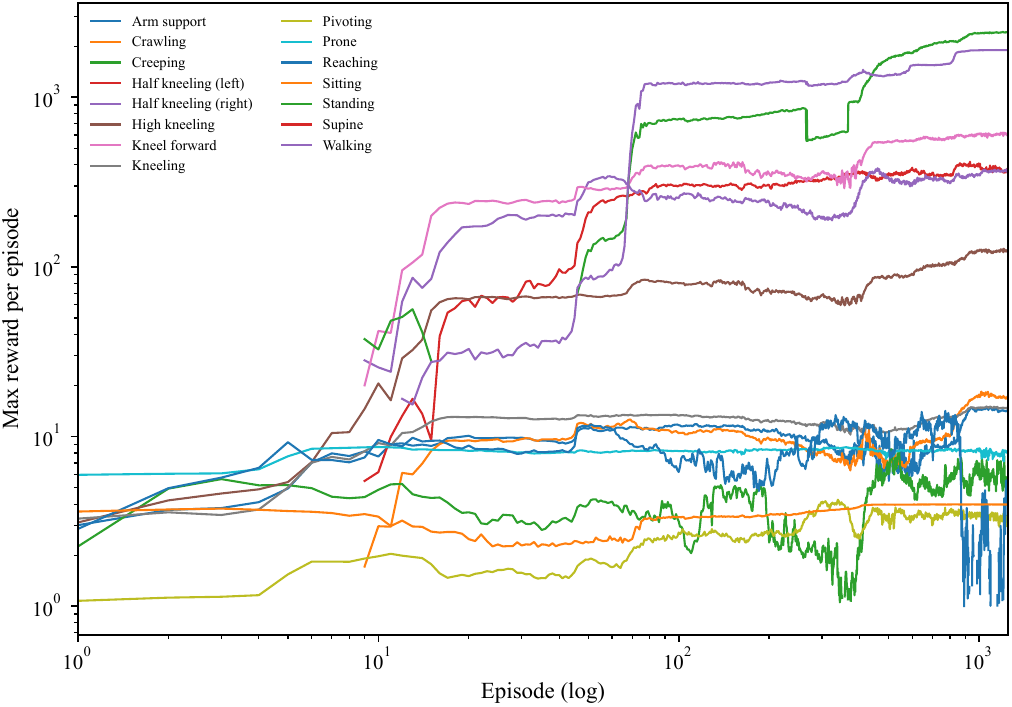}
    \caption{Total reward for each milestone during training using the GMFM graph with a 50\% passing score. Rewards are shown in log scale, suggesting that more advanced milestones dominate learning in later episodes.}
    \label{reward_trajectories}
\end{figure*}

\begin{figure}[tbp]
     \smallskip
     \smallskip
     \smallskip
    \centering
    \includegraphics[width=0.48\textwidth]{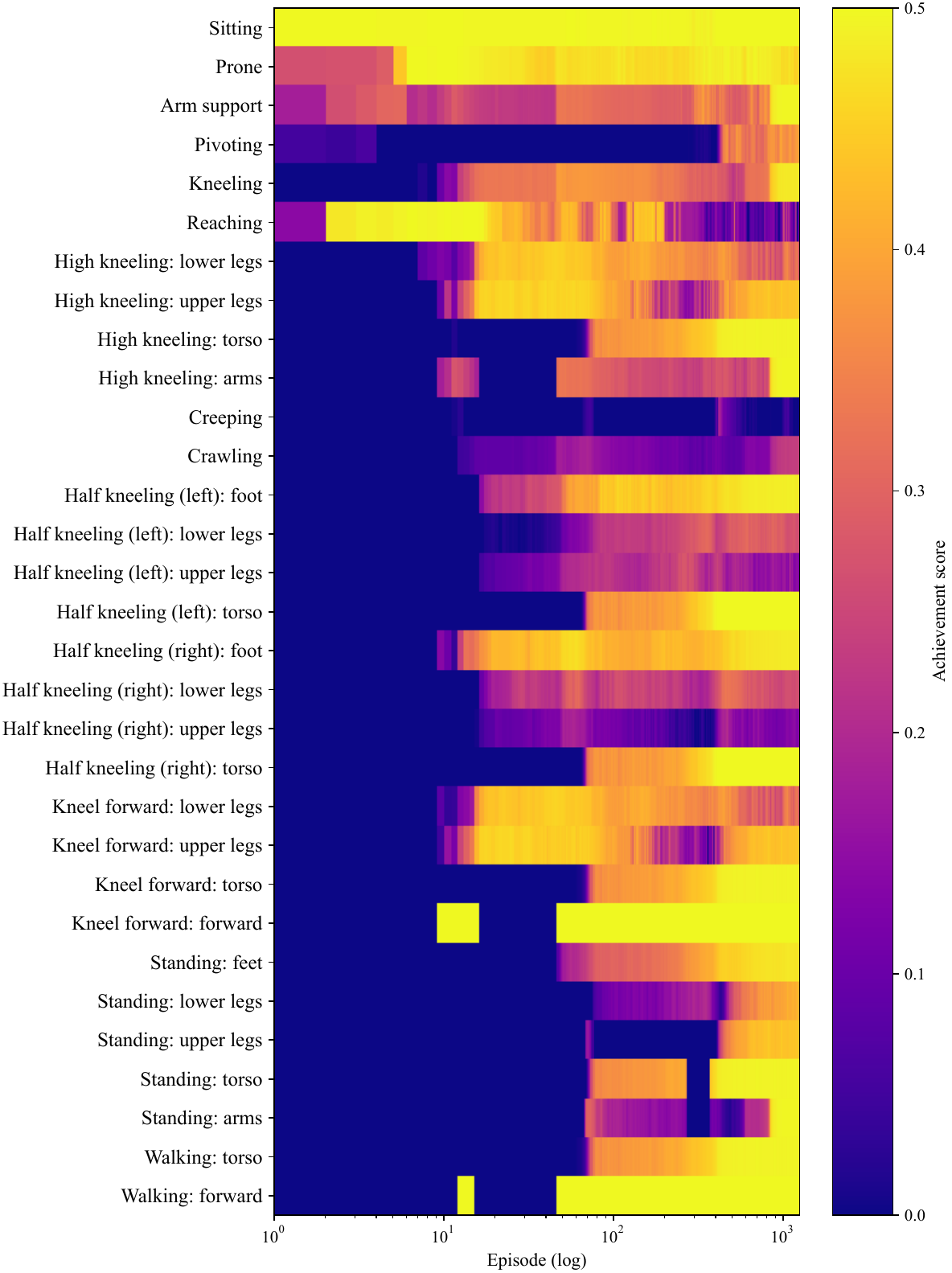}
    \caption{Achievement-score heatmap for each milestone and individual component during training using the GMFM graph with a 50\% passing score. Alternating dark and light regions indicate that different milestones are optimized at different learning stages.}
    \label{score_heatmap}
\end{figure}

Fig.~\ref{reward_trajectories} and Fig.~\ref{score_heatmap} illustrate the progression of rewards and achievement scores for each milestone during typical training episodes. The results suggest that the presence of a kneeling-to-standing bridge is an important structural feature for learning upright locomotion from the ground up. The supplemental video further shows that the kneel-forward milestone in the GMFM graph allows the robot to learn an efficient alternating gait even though no explicit reward or regularization term was provided for gait quality.

\begin{table}[tbp]
     \smallskip
     \smallskip
     \smallskip
\centering
\caption{Stand success rate and walking speed for different graphs and passing scores.}
\label{tab:success_rates}
\renewcommand{\arraystretch}{1.15}
\begin{tabular}{cccc}
\toprule
\textbf{Graph} & \textbf{\makecell{Passing \\ score}} & \textbf{\makecell{Stand \\ success rate}} & \textbf{\makecell{Walking speed \\ (m/s)}} \\
\midrule
\multirow{4}{*}{AIMS} & 50\% & 99\% & 0.87 \\
                    & 60\% & 99\% & 0.96 \\
                    & 70\% & 98\% & 0.92 \\
                    & 80\% & 90\% & 0.99 \\
\midrule
\multirow{4}{*}{BSID} & 50\% & 100\% & 0.95 \\
                    & 60\% & 99\% & 1.12 \\
                    & 70\% & 99\% & 1.34 \\
                    & 80\% & 93\% & 0.99 \\
\midrule
\multirow{4}{*}{GMFM} & 50\% & 99\% & 1.33 \\
                    & 60\% & 100\% & 1.44 \\
                    & 70\% & 100\% & 1.52 \\
                    & 80\% & 99\% & 1.02 \\
\midrule
\multirow{4}{*}{PDMS} & 50\% & 0\% & -- \\
                    & 60\% & 99\% & 0.95 \\
                    & 70\% & 99\% & 0.45 \\
                    & 80\% & 0\% & -- \\
\bottomrule
\end{tabular}
\end{table}

\begin{table*}[tbp]
\centering
\begin{threeparttable}
\caption{Episode when each milestone is first achieved for different graphs and passing scores.}
\label{tab:graph_comparison}
\renewcommand{\arraystretch}{1.15}
\begin{tabular}{lrrrrrrrrrrrrrrrr}
\toprule
& \multicolumn{4}{c}{\textbf{AIMS}} & \multicolumn{4}{c}{\textbf{BSID}} & \multicolumn{4}{c}{\textbf{GMFM}} & \multicolumn{4}{c}{\textbf{PDMS}} \\
\cmidrule(lr){2-5} \cmidrule(lr){6-9} \cmidrule(lr){10-13} \cmidrule(lr){14-17}
\textbf{Passing Score} & 50\% & 60\% & 70\% & 80\% & 50\% & 60\% & 70\% & 80\% & 50\% & 60\% & 70\% & 80\% & 50\% & 60\% & 70\% & 80\% \\
\midrule
Prone & 0 & 1 & 1 & 1 & 0 & 0 & 1 & 1 & 0 & 0 & 1 & 1 & 1 & 1 & 1 &  \\
Sitting & 0 & 0 & 0 & 0 & 0 & 0 & 0 & 0 & 0 & 0 & 0 & 1 & 0 & 0 & 0 & 0 \\
Scooting & -- & -- & -- & -- & -- & -- & -- & -- & -- & -- & -- & -- & 0 & 0 & 0 & 1 \\
Arm support & 1 & 2 & 2 & 3 & 1 & 1 & 2 & 6 & 1 & 1 & 2 & 7 & 1 & 1 & 2 &  \\
Reaching & 1 & 2 & 2 & 4 & 1 & 1 & 2 & 8 & 1 & 1 & 2 & 7 & 1 & 2 &  &  \\
Pivoting & 1 & 3 & 37 & 113 & -- & -- & -- & -- & 1 & 54 & 3 & 70 & -- & -- & -- & -- \\
Kneeling & 6 & 5 & 5 & 9 & 4 & 9 & 7 & 11 & 7 & 5 & 7 & 5 & 7 & 76 & 356 &  \\
High kneeling & -- & -- & -- & -- & -- & -- & -- & -- & 7 & 5 & 7 & 4 & -- & -- & -- & -- \\
Half kneeling (L) & 12 & 37 & 18 & 74 & 4 & 10 & 9 & 20 & 16 & 20 & 13 & 71 & -- & -- & -- & -- \\
Half kneeling (R) & 30 & 32 &  &  & 15 & 26 & 16 & 16 & 9 & 13 & 22 & 165 & -- & -- & -- & -- \\
Kneeling (L) foot & 11 & 12 & 8 & 15 & -- & -- & -- & -- & -- & -- & -- & -- & -- & -- & -- & -- \\
Kneeling (R) foot & 14 & 12 & 13 & 18 & -- & -- & -- & -- & -- & -- & -- & -- & -- & -- & -- & -- \\
Kneel forward & -- & -- & -- & -- & -- & -- & -- & -- & 9 & 13 & 10 & 26 & -- & -- & -- & -- \\
Crawling & 9 & 27 & 47 &  & 9 & 12 & 200 & 323 & 12 & 16 & 17 & 26 & 585 & 961 & 374 &  \\
Creeping & 14 & 67 &  &  & 19 & 14 & 211 & 301 & 11 & 39 & 43 & 32 &  & 174 & 359 &  \\
Quadruped & 15 & 116 & 27 & 93 & -- & -- & -- & -- & -- & -- & -- & -- & -- & -- & -- & -- \\
Squat & 20 & 116 & 27 & 93 & -- & -- & -- & -- & -- & -- & -- & -- & -- & -- & -- & -- \\
Standing & 19 & 116 & 28 & 103 & 18 & 18 & 400 & 424 & 46 & 44 & 26 & 166 &  & 1101 & 354 &  \\
Walking & 9 & 109 & 28 & 100 & 13 & 12 & 295 & 375 & 12 & 20 & 17 & 59 &  & 205 & 381 &  \\
\bottomrule
\end{tabular}
\begin{tablenotes}
    \small
    \item Empty cells indicate that the milestone was not achieved in that run. ``--'' indicates that the milestone was not present in the scale.
\end{tablenotes}
\end{threeparttable}
\end{table*}

\begin{table}[tbp]
     \smallskip
     \smallskip
     \smallskip
\centering
\begin{threeparttable}
\caption{Time to milestone achievement (seconds).}
\label{tab:milestone_times}
\begin{tabular}{lccccc}
\toprule
Milestone & $p_5$ & $p_{25}$ & $p_{50}$ & $p_{75}$ & $p_{95}$ \\
\midrule
Sitting & 1.1 & 1.2 & 1.2 & 1.3 & 1.3 \\
Prone & 1.5 & 1.5 & 1.6 & 1.7 & 1.7 \\
Arm support & 1.5 & 1.5 & 1.6 & 1.7 & 1.9 \\
Pivoting & 3.8 & 4.6 & 5.6 & 6.4 & 7.7 \\
Kneeling & 1.8 & 2 & 2.6 & 2.8 & 2.9 \\
Reaching & 2.2 & 2.9 &  &  &  \\
High kneeling: lower legs & 1.7 & 1.7 & 1.8 & 1.9 & 2 \\
High kneeling: upper legs & 2.3 & 2.4 & 2.5 & 2.6 & 2.8 \\
High kneeling: torso & 2.8 & 2.9 & 3.0 & 3.1 & 3.3 \\
High kneeling: arms & 2.8 & 2.9 & 3.0 & 3.1 & 3.3 \\
Creeping & 3.5 &  &  &  &  \\
Crawling & 3.0 & 3.1 & 3.2 & 3.3 & 3.5 \\
Half kneeling (L): foot & 3.1 & 3.3 & 3.4 & 3.5 & 3.7 \\
Half kneeling (L): lower legs & 3.2 & 3.3 & 3.4 & 3.5 & 3.7 \\
Half kneeling (L): upper legs & 3.2 & 3.3 & 3.4 & 3.5 & 3.7 \\
Half kneeling (L): torso & 3.2 & 3.3 & 3.4 & 3.5 & 3.7 \\
Half kneeling (R): foot & 3.0 & 3.1 & 3.2 & 3.3 & 3.5 \\
Half kneeling (R): lower legs & 3.1 & 3.3 & 3.3 & 3.5 & 3.7 \\
Half kneeling (R): upper legs & 3.1 & 3.3 & 3.4 & 3.5 & 3.7 \\
Half kneeling (R): torso & 3.1 & 3.3 & 3.4 & 3.5 & 3.8 \\
Kneel forward: lower legs & 3 & 3.1 & 3.2 & 3.3 & 3.5 \\
Kneel forward: upper legs & 3 & 3.1 & 3.2 & 3.3 & 3.5 \\
Kneel forward: torso & 3 & 3.1 & 3.2 & 3.3 & 3.5 \\
Kneel forward: forward & 3.1 & 3.2 & 3.3 & 3.4 & 3.7 \\
Standing: feet & 3.1 & 3.2 & 3.3 & 3.4 & 3.6 \\
Standing: lower legs & 3.4 & 3.5 & 3.6 & 3.7 & 3.9 \\
Standing: upper legs & 3.4 & 3.5 & 3.6 & 3.7 & 3.9 \\
Standing: torso & 3.4 & 3.5 & 3.6 & 3.7 & 3.9 \\
Standing: arms & 3.4 & 3.5 & 3.6 & 3.7 & 3.9 \\
Walking: torso & 3.0 & 3.1 & 3.2 & 3.3 & 3.5 \\
Walking: forward & 3.1 & 3.2 & 3.3 & 3.4 & 3.6 \\
\bottomrule
\end{tabular}
\begin{tablenotes}
    \small
    \item Data obtained by executing the trained GMFM policy with a 50\% passing score on a randomized population of robots.
\end{tablenotes}
\end{threeparttable}
\end{table}

\addtolength{\textheight}{-1.0cm}   

Table~\ref{tab:milestone_times} shows the 5th--95th percentiles of time to milestone achievement when executing a trained model on a large population of robots with randomized physical and mechanical properties. The small variance shows that the developmental-graph approach allows the robot to achieve milestones at roughly the same time across all robots, suggesting that the learned policy can systematically perform a chain of actions rather than relying on random exploration.

\section{Conclusions}

We showed that developmental graphs derived from human pediatric motor scales can guide a simulated humanoid robot to stand up and walk without reference trajectories (e.g., from learning by demonstration) and with no need for hand-tuning reward weights. Three of the four scales we tested succeeded under all four tested training settings. The fourth scale failed at the lowest and highest tested passing scores, and a targeted test in which we inserted the missing half-kneeling milestone shows that this structural gap is a likely cause of the failure. This validates both the feasibility of using clinical developmental scales for humans as robot curricula and the specific milestone structure they prescribe.

This framework is novel and practical in that it leverages the human motor developmental scales to enable systematic robot motor skill learning rather than the prevailing literature of robot RL through large and random exploration of motion data guided by human demonstration. As such, this framework does not require ad hoc tuning of parameters, enables automatic determination of reward weight assignment from graph levels, and uses decomposition of complex milestones into force-propagation mini-graphs. Future work can examine other multiplier growth-rate schemes, combine the training setup with robot morphology changes, extend this approach to finer motor skills, and transfer it to the real world.

\bibliography{ref}

\end{document}